\documentclass[runningheads]{llncs}

\usepackage{accv}

\usepackage{accvabbrv}

\usepackage{graphicx}
\usepackage{booktabs}
\usepackage{multirow}
\usepackage{color, colortbl}
\definecolor{Gray}{gray}{0.9}
\definecolor{lightgreen}{rgb}{0.56, 0.93, 0.56}
\definecolor{lightskyblue}{rgb}{0.53, 0.81, 0.98}
\usepackage[accsupp]{axessibility}  

\usepackage[pagebackref,breaklinks,colorlinks,citecolor=accvblue]{hyperref}

\usepackage{orcidlink}

\begin{document}

\title{GLMHA --- A Guided Low-rank Multi-Head Self-Attention for Efficient Image Restoration and Spectral Reconstruction} 

\titlerunning{GLMHA}

\author{Zaid Ilyas\inst{1} \and Naveed Akhtar\inst{2} \and David Suter\inst{1} \and
Syed Zulqarnain Gilani\inst{1}}

\authorrunning{Z.~Ilyas et al.}

\institute{Centre for AI \& ML, School of Science, Edith Cowan University, Australia
\email{z.ilyas@ecu.edu.au}\\
School of Computing and Information Systems
The University of Melbourne, Melbourne, Australia}

\maketitle

\begin{abstract}
Image restoration and spectral reconstruction are longstanding computer vision tasks. Currently, CNN-transformer hybrid models provide state-of-the-art performance for these tasks. The key common ingredient in the architectural designs of these models is Channel-wise Self-Attention (CSA). We first show that CSA is an overall low-rank operation. Then, we propose an instance-Guided Low-rank Multi-Head self-attention (GLMHA) to replace the CSA for a considerable computational gain while closely retaining the original model performance. Unique to the proposed GLMHA is its ability to provide computational gain for both short and long input sequences. In particular, the gain is in terms of both Floating Point Operations (FLOPs) and parameter count reduction. This is in contrast to the existing popular computational complexity reduction techniques, e.g., Linformer, Performer, and Reformer, for whom FLOPs overpower the efficient design tricks for the shorter input sequences. Moreover, parameter reduction remains unaccounted for in the existing methods. We perform an extensive evaluation for the tasks of spectral reconstruction from RGB images, spectral reconstruction from snapshot compressive imaging, motion deblurring, and image deraining by enhancing the best-performing models with our GLMHA. Our results show up to a 7.7 Giga FLOPs reduction with 370K fewer parameters required to closely retain the original performance of the best-performing models that employ CSA. 
\end{abstract}

\section{Introduction}
\label{sec:intro}

Image restoration and spectral reconstruction are key image-processing tasks~\cite{r22,r23,r24,r35,r36,r37,r41}. Both recover missing or degraded information in images. The former achieves this along the spatial dimensions of the image~\cite{r22,r24,r23}, whereas the later aims at the same along the spectral dimension~\cite{r36,akhtar2018hyperspectral,r38,r39,r41}.
Currently, CNN-transformer hybrid neural models, e.g.,~\cite{r1,r42,r43} establish the state-of-the-art (SOTA) for both these tasks. 

Owing to the intrinsic similarities between the  objectives of image restoration and spectral reconstruction, the techniques achieving the SOTA performance for these tasks exhibit certain similarities in the underlying neural networks. 
The key common ingredient of these neural networks is the Channel-wise Self-Attention (CSA) in the transformer blocks that handles images with different resolutions.
The CSA deals with channels of an input feature map as  sequence samples and hence computes self-attention among the \textit{channels} rather than the spatial elements or spatial patches, which is done in a typical vision transformer model, e.g., ViT \cite{r44}, Swin Transformer \cite{r45}.



\begin{figure}[t]
\begin{center}
\includegraphics[scale =0.55]{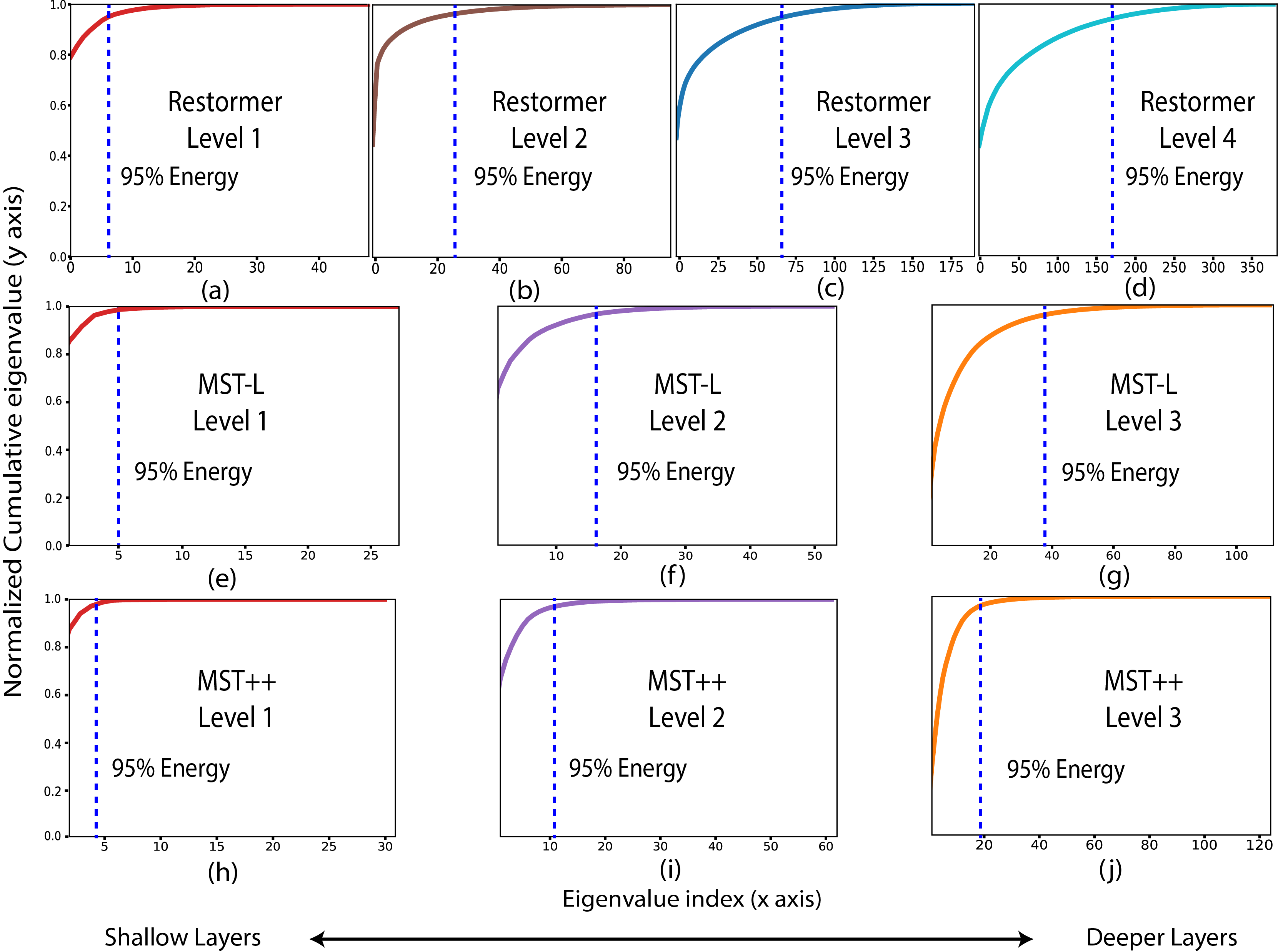}
\end{center}
   \caption{Low-rank nature of self-attention. Average cumulative distribution function of Eigenvalues for self-attention of Restormer~\cite{r1} (a-d), MST-L~\cite{r42} (e-g), and MST++~\cite{r43} (h-j). For Restormer, the plot averages image deraining, denoising, and motion deblurring performance. For MST-L, we perform real and simulated data spectral reconstruction. For MST++, NTIRE 2022 challenge dataset for spectral reconstruction is used. Further details with more supporting results are also provided in supplementary material.   
   }
\label{fig_2}
\end{figure}

In \cref{fig_2}, we empirically establish  that most of the signal's energy for CSA is generally retained by a small fraction of Eigenvalues, especially in the early layers of the networks. Our Singular Value Decomposition analysis identifies  the possibility of a more efficient low-rank approximation of the CSA operation, which is the eventual key contribution of this paper. We note that, in general, reducing the computational cost of transformer-based neural models is an active research direction. In fact, it is possible to broadly categorize the contemporary techniques in this direction into sparsity-based \cite{rs1,r3,r4,r5,r6,r7,r14} and low-rank approximation techniques \cite{r_new_1, ,r9,r12,r13}.  The former methods focus on attending to a subset of input sequence, whereas the later first generate same-sized query, key, and value embeddings; as in the case of conventional self-attention, and then perform optimization steps like low-rank projections \cite{r2}, Locality Sensitive Hashing (LSH) \cite{r8} etc., to reduce the $\mathcal{O}(n^{2})$ complexity of the attention matrix calculation.


Though effective for longer input sequences, utilizing these techniques for shorter sequences, which is required for CSA, severely limits their effectiveness. {Based on our experiments we found that} for the shorter sequences, the computational overhead of the optimization overpowers the arithmetic complexity reduction.
Moreover, the methods also fail in reducing the number of parameters for the self-attention calculation. Indeed, there are other methods \cite{r42,r28,r4,r64} that target both computational cost and memory footprint reduction. However, they achieve it by reducing the length of the input sequence using downsampling, or by limiting the scope for self-attention, or employing a higher number of self-attention heads. All these adversely affect the model performance due to information loss and/or inadequate long-range dependency. 

Considering the ultimate objective of low-rank self-attention, one obvious approach to parameter and FLOPs reduction is the direct generation of low-rank key \underbar{\textit{K}} and value \underbar{\textit{V}} matrices from the input feature map \textit{X}. However, this is also known to result in unacceptable  approximation errors~\cite{r_bottleneck}. Owing to the inherent low-rank nature of self-attention - see \cref{fig_2}, it can be expected that for an input instance, there are key/value candidate pairs that regardless of their location in the sequence, might hold a relatively higher semantic importance to the query. Hence, we argue that the low-rank embeddings generation can benefit greatly from instance-guidance to maximizes the important information content. 

In this work, we introduce an instance-Guided Low-rank Multi-Head self-Attention (GLMHA).
Our technique leverages a light  calibration network that takes query embedding 
as input and generates calibration weights that modify the input feature map that is subsequently used to generate the relevant information-rich low-rank key/value embeddings. The proposed method has three major unique benefits. (\textit{i}) It leads to both parameter count and computation cost reduction for self-attention. (\textit{ii}) It is amicable to small input sequences, unlike the existing methods. (\textit{iii}) It closely retains the original model performance despite achieving much larger parameter count and FLOPs reduction as compared to the existing methods. We extensively evaluate the abilities of GLMHA by comparing its performance against three popular complexity reduction methods Linformer~\cite{r2}, Reformer~\cite{r8}, and Performer~\cite{r9} for the tasks of motion deblurring, image deraining, hyperspectral reconstruction from color images, and hyperspectral reconstruction from snapshot compressive imaging. Our method achieves up to 7.7 Giga FLOPs reduction and 0.37 Million parameter reduction in the base models. For a thorough analysis, our experiments are presented using three mainstream base models for the tasks, i.e., Restormer \cite{r1}, MST-L \cite{r42}, and MST++ \cite{r43}.


\section{Related Work}
Our contribution encompasses three directions of image restoration, spectral reconstruction and efficient transformer architectures. Hence, we review advances in these directions in this section.

 \noindent\textbf{Image restoration:} Initial efforts in  image restoration include traditional approaches \cite{r20,r21,r22,r24,r23}, which were later outperformed by the Convolutional Neural Network (CNN) based methods~\cite{r15,r16,r17,r18,r19}. These techniques mainly rely on the encoder-decoder paradigm for image restoration. However, in general, CNN architectures lack long-range dependency considerations for image pixels. This compromises their image restoration abilities. More recently, employing attention mechanism in neural models for image restoration is gaining popularity~\cite{r25,r26,r27}.  
 In particular, transformer-based models currently achieve excellent performance for this task~\cite{r1,r28,r29}.  Among these methods, Restormer \cite{r1} is the best performer which is a hybrid CNN-transformer model  that leverages CSA in its  transformer blocks. It has the lowest FLOPs and parameter count, which is particularly attractive.\par 
 \noindent\textbf{Spectral image reconstruction:} Early attempts for spectral image reconstruction  mainly leveraged handcrafted priors \cite{r30,r31,r32,r33,r34,r35, akhtar2018hyperspectral}. The limited representation capacity of such techniques leads to inadequate performance for this task. The more recent CNN-based techniques outperformed the traditional methods due to their stronger representation prowess  \cite{r36,r37,r38,r39,r40,r41}.  Nevertheless, these approaches also suffer from the inadequacy of  capturing the long-range inter-dependencies and non-local self-similarities. More recently, transformer-based methods, MST \cite{r42} and MST++ \cite{r43} report state-of-the-art performance  for this task. These methods also employ {CSA} mechanism. Hence, they are the primary beneficiary of our contribution.  

\noindent\textbf{Efficient transformers:}
Existing efforts to make transformers computationally efficient, can be broadly categorized into two types, namely; Sparse Attention techniques \cite{rs1,r3,r4,r5,r6,r7,r14} and Low-Rank approximation methods \cite{r2,r9,r12,r13}. The Sparse Attention paradigm reduces the computational cost of the models by only attending to a subset of the input sequence, rather than all possible pairs. 
For instance, Reformer \cite{r8} achieves sparsity by using Locality Sensitive Hashing (LSH), which allows the model to only attend to a subset of the input tokens. It reduces the complexity of self-attention calculation from $\mathcal{O}(n^{2})$ to $\mathcal{O}(n$log$n)$. 
Image Transformer \cite{r4} and Block-wise Self Attention \cite{r3} use block-wise self-attention for this purpose. However, they enable only local attention. To address that, approaches like Sparse Transformer \cite{r5}, Big Bird \cite{r6} and Longformer \cite{r7} use a combination of block-wise and dilated window strategies to capture both local and global context for attention. These techniques lower the computational cost but fail to reduce the model parameter count.
Low-rank approximation methods~\cite{r2,r9,r12,r13} transform the attention matrix calculation into a low-rank operation. Linformer \cite{r2} and Performer \cite{r9} are prominent examples of this direction. Linformer \cite{r2} generates low-rank Key and Value embeddings and then applies self-attention which reduces the computational cost to $\mathcal{O}(n)$. Performer \cite{r9} uses orthogonal random features to perform a low-rank approximation of the self-attention matrix. There are other works that aim at combining both sparse and low-rank approximations, e.g., ScatterBrain \cite{r10}. However, all these methods must rely on longer sequences, which makes them ineffective for the tasks like image restoration and spectral reconstruction. 

\section{Proposed Method}
In this work, we introduce an instance-Guided Low-rank Multi-Head self-Attention (GLMHA) mechanism that is particularly suited to the critical low-level image processing tasks of restoration and spectral reconstruction. To comprehend the proposed technique,  it is imperative to review the notion of Channel-wise Selt-Attention (CSA). Hence, we first provide a brief discussion on the key concept of CSA before delving into the details of GLMHA.

\subsection{Channel-wise Self-Attention (CSA)}
\label{sec:CSA}
\begin{figure}[t]
\begin{center}
\includegraphics[width = 0.7\textwidth]{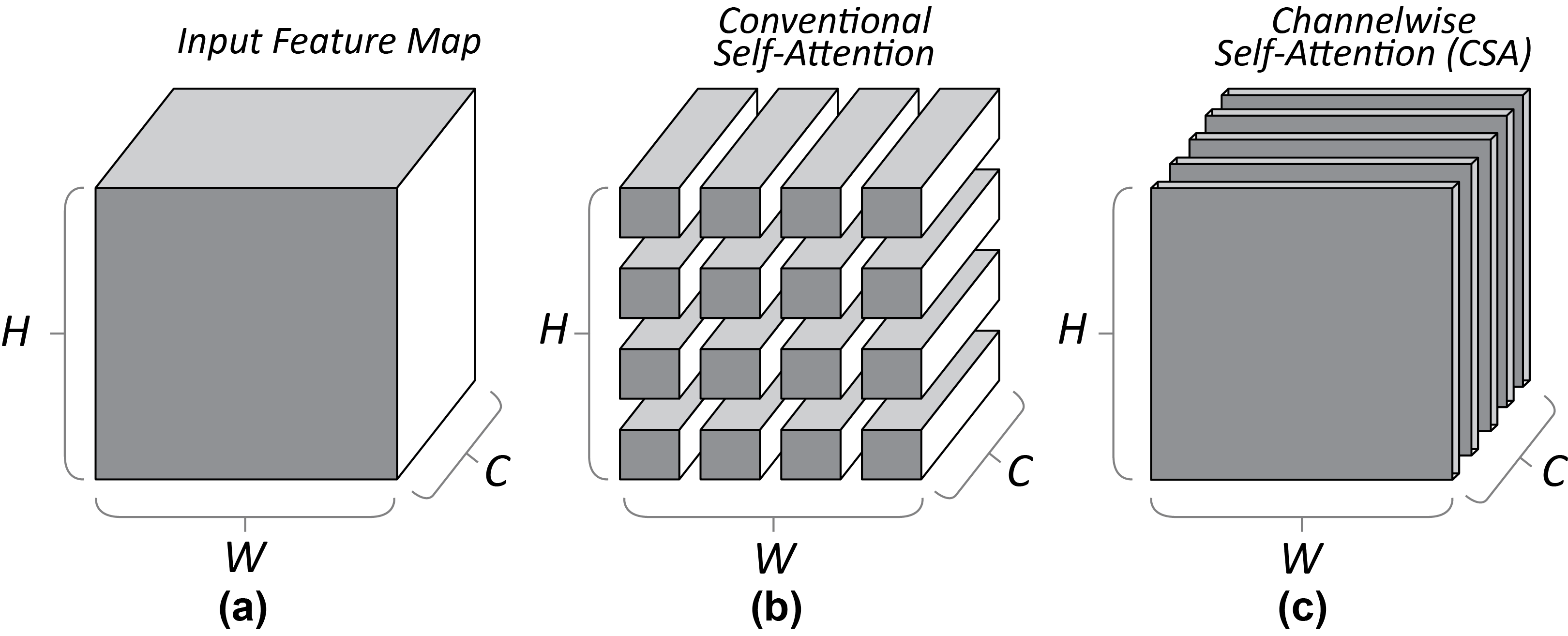}
\end{center}

   \caption{(a) Input Feature Map. (b) The conventional self-attention approach operates on samples or patches in \textit{HW} domain. (c) Channel Self-Attention (CSA) operates on channels \textit{C},  treating each channel as a sample in the sequence and \textit{HW} as its embedding. }
\label{fig_1}
\end{figure}

Currently, the existing state-of-the-art models for image restoration and spectral image reconstruction, e.g., Burstformer \cite{r_csa_5}, PromptIR \cite{r_csa_4}, MST \cite{r42}, MST++ \cite{r43}, and Restormer \cite{r1} employ Channel-wise Self-Attention (CSA). As illustrated in \cref{fig_1}, CSA operates differently from the widely used conventional self-attention. 
It focuses on leveraging correlation along the channels of the input feature map instead of the spatial correlations. This, in part, is motivated by the goal of handling inputs with varying resolutions without a quadratic increase in the computational cost with the input size increase.
Given an input feature map \textit{X}~$\in \mathbb{R}^{H\times W\times C}$, the conventional self-attention must compute a large attention matrix of size $\mathbb{R}^{HW\times HW}$. In contrast, CSA
first calculates Queries (\textit{Q}), Keys (\textit{K}), and Values (\textit{V}) projections of size $\mathbb{R}^{C\times HW}$. Then, it  reshapes the \textit{Q} and \textit{K} projections to get a dot product of size $\mathbb{R}^{C\times C}$.  The resulting efficient self-attention process is formulated as
\begin{equation}
\textit{Q} = W_{Q}\textit{X},~~\textit{K} = W_{K}\textit{X},~~\textit{V} = W_{V}\textit{X},
\end{equation}
\begin{equation}
    \textit{Z} = \mbox{Softmax}(\textit{Q} \cdot \textit{K}^{\intercal}/\beta) \cdot \textit{V} + \textit{X},
\end{equation}
where \textit{X} and \textit{Z} are the input and output features of the self-attention layer. $W_{Q}$, $W_{K}$ and $W_{V}$ are the weight matrices to calculate Queries (\textit{Q}), Keys (\textit{K}), and Values (\textit{V}) projections. $\beta$ is a learnable scaling parameter that controls the intensity of \textit{Q} and \textit{K} dot product.

\begin{figure*}[t]
\begin{center}
\includegraphics[width =\textwidth]{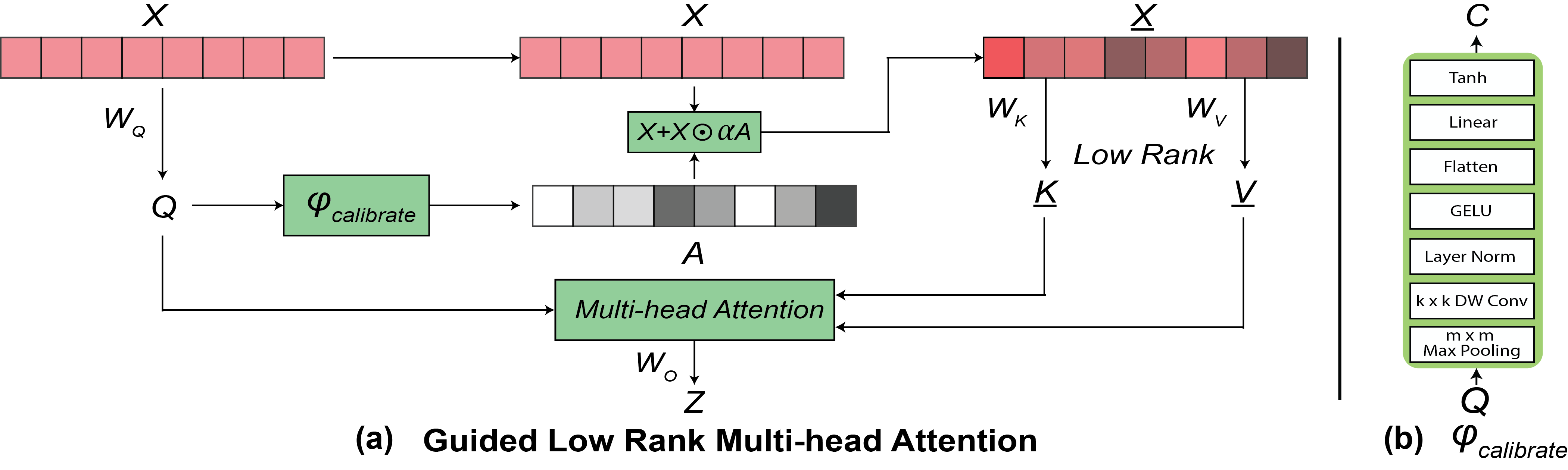}
\end{center}
   \caption{(a) Pipeline of the proposed GLMHA. (b) Calibration Network}

\label{fig_3}
\end{figure*}

\subsection{Guided Low-rank Multi-Head self-Attention}
\label{sec:GLMHA}
The proposed Guided Low-rank Multi-Head self-Attention (GLMHA) is devised to boost the efficiency of Channel-wise Self-Attention (CSA). Although CSA is already tailored to provide computational gains over the conventional attention, techniques employing CSA, e.g., MST \cite{r42}, MST++ \cite{r43}, Restormer \cite{r1}, still demand very  heavy computations because of the nature of the  image restoration and spectral reconstruction tasks.
The proposed GLMHA is an important step towards  lowering these computational requirements while minimally affecting the task performance.

Consider a sequence of length $n$. We already show in \cref{fig_2} that CSA has a low-rank nature. Hence, it is imperative that for any query $Q$, there are $m: m<n$ samples in the Key $K$ embeddings that would hold a high correlation to the query, while the remaining $n$ - $m$ samples can be neglected with minimal loss of information. Moreover, it is plausible to improve the approximation by further amplifying the contribution of the relevant samples in the input feature map \textit{X} and reducing the impact of less relevant features. This is a key guiding principle for GLMHA. We illustrate the schematics of GLMHA in \cref{fig_3}.  

\noindent\textbf{Calibration Vector Generation ---} 
Given an input feature map \textit{X} $\in$ $\mathbb{R}^{C \times HW}$, where \textit{C}, \textit{H}, and \textit{W} is the number of channels, height, and width of the  map respectively, we generate Query embedding \textit{Q}  from \textit{X} using a linear transformation, i.e., $Q = W_{Q}X$. The generated \textit{Q} embedding is reshaped to $\mathbb{R}^{H \times W \times C}$ and is fed to a calibration network $\varphi_{calibrate}$. The eventual goal of this sub-network is to 
generate an instance-based weighting vector \textit{A} that provides more weights to the samples that are semantically close to the input sequence \textit{X}, and does the opposite to the less relevant samples, regardless of their position in the sequence. 

The sub-network $\varphi_{calibrate}$ performs max pooling operation to compress spatial information in $Q$. This is followed by a lightweight depth-wise convolution operation with GELU activation that extracts useful spatial information. Finally, a linear layer with \textit{tanh} activation compresses the information and generates the calibration vector \textit{A}. The sub-network is illustrated in \cref{fig_3}(b). We summarize this $X \rightarrow A$ transformation as
\begin{equation}
\textit{Q} = W_{Q}\textit{X},
\end{equation}
\begin{equation}
\textit{A} = \varphi_{calibrate}(\textit{Q}).
\end{equation}

The coefficients of $A$ encode  different levels of importance for different samples of the input.
We further employ a hyperparameter $\alpha$ as a scaling factor for \textit{A} to control the influence of its coefficients. Let \textit{\underbar{X}}$\in \mathbb{R}^{C \times HW}$ be the re-weighted \textit{X}. We eventually compute it as follows.
\begin{equation}
    \underbar{\textit{X}} = \textit{X}+ (\textit{X}\odot\alpha\textit{A}),
\end{equation}
where $\odot$ is the Hadamard product. The objective of this manipulation is that when $\alpha$\textit{A} $= 0$,  unit weight is assigned to the sample of the input sequence. For $\alpha$\textit{A} $> 0$   a proportionally larger weight is used and, for $\alpha$\textit{A} $< 0$, proportionally smaller re-weighting occurs. Adding   \textit{X}$\odot \alpha A$ back to \textit{X} also provides numerical stability.
Specific to instance \textit{Q}, the resultant feature map \textit{\underbar{X}} achieves enhanced values for the samples that are semantically closer and more relevant to the input instance, and diminished values for the less important samples.\par




\noindent\textbf{Low-Rank Self-Attention ---} 
We employ the resultant feature map \textit{\underbar{X}} $\in$ $\mathbb{R}^{C \times HW}$  to generate low-rank Key and Value embeddings directly in a learnable manner, which helps in  saving the parameters and FLOPs. Moreover, the generated embeddings are rich in semantically relevant information. In our case, given \textit{\underbar{X}} $\in$ $\mathbb{R}^{C \times HW}$,
\begin{equation}
\textit{\underbar{K}} = W_{\textit{\underbar{K}}}\underbar{\textit{X}}, ~~~~~
\textit{\underbar{V}} = W_{\textit{\underbar{V}}}\underbar{\textit{X}} ,
\end{equation}
\begin{equation}
    \textit{Z} = \mbox{Softmax}(\textit{Q} \cdot \textit{\underbar{K}}^{\intercal}/\beta) \cdot \textit{\underbar{V}} + \textit{X},
\end{equation}
where \textit{\underbar{K}} $\in$ $\mathbb{R}^{C' \times HW}$ and \textit{\underbar{V}} $\in$ $\mathbb{R}^{C' \times HW}$ are the generated low-rank embeddings, where $C' < C$. The difference between $C'$ and $C$ is that the former is controllable with a predefined percentage of size reduction for the low-rank matrices. $\beta$ is a learnable scaling parameter that controls the \textit{Q} and \textit{\underbar{K}} dot product. We enventually implement GLMHA as a multi-head attention layer. 
\textit{X} and \textit{Z} are respectively the input and the output of the GLMHA layer with shape $\mathbb{R}^{H \times W \times C}$. \par 

\begin{figure*}[t]
\begin{center}
\includegraphics[width = 1\textwidth]{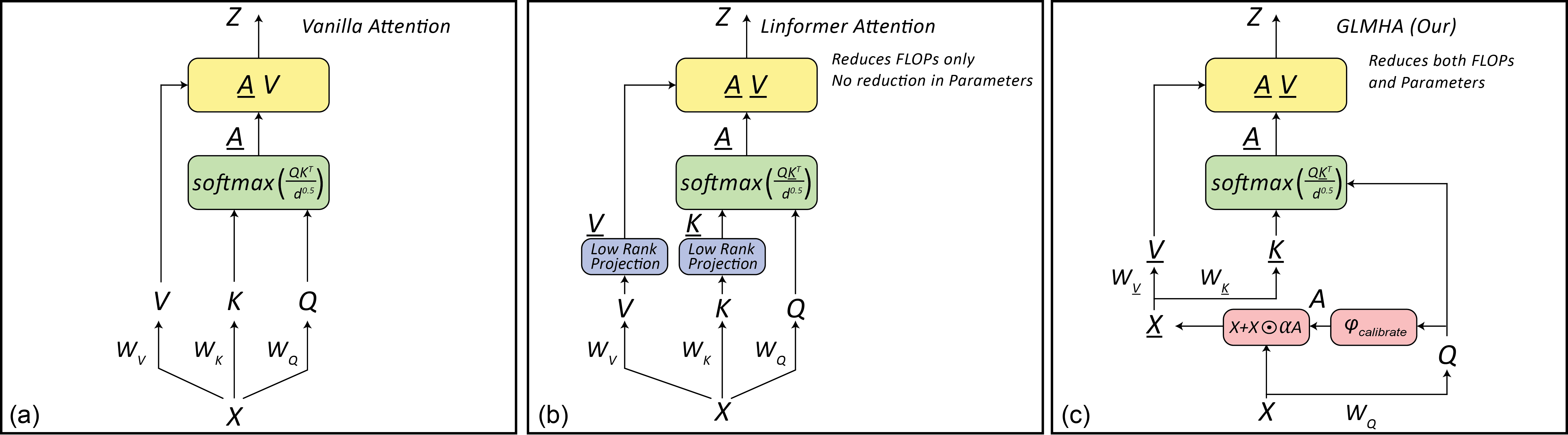}
\end{center}
   \caption{(a) Conventional self-attention (b) Linformer low-rank approximation-based self-attention. The generated Keys and Values are projected to low-rank approximations and then self-attention is calculated. (c) Proposed GLMHA that generates low-rank Keys and Values embeddings from input feature map in an instance-based learnable way.}

\label{fig_4}
\end{figure*}

\noindent\textbf{Computational Complexity ---}
The GLMHA has a linear computational complexity of $\mathcal{O}(n)$ with parameter count reduction ratio with respect to the full-scaled self-attention of
\begin{equation}
\frac{\left (  2n^{2}-\frac{n^{2}}{h^{2}}-nk^{2}\right)r-2n^{2}}{3n^{2}r},
\label{eq:factor}
\end{equation}
where \textit{n} is the sequence length, i.e., full-sized length (number of channels) of \textit{K} and \textit{V}, \textit{h} is the number of heads, \textit{r} is the reduction factor for \textit{\underbar{K}} and \textit{\underbar{V}} generation, and $k^{2}$ is the kernel size of the underlying depth-wise convolution operation. We provide derivation for the expression in \cref{eq:factor} in the supplementary material of the paper. It can be observed that GLMHA also becomes  more effective in saving parameters when the  sequence length \textit{n} (number of channels in the case of CSA) is large. 
This is a highly desirable property for a computational complexity reduction method that is also effective for shorter sequences.  

\noindent\textbf{Difference from Linformer ---} 
Linformer~\cite{r2} is a popular method that has close relevance to GLMHA. Hence, it is imperative to highlight the difference between the two. Our  GLMHA consists of an instance-guided weighting vector generation mechanism and low-rank multi-head self-attention. It is different from Linformer in  that the Linformer first generates \textit{Q}, \textit{K}, and \textit{V} embeddings and then finds low-rank approximations, say  \textit{\underbar{K}} and \textit{\underbar{V}}, to calculate low-rank self-attention. Comparatively, this adds additional FLOPs and there is no reduction in the parameter count. Although the adverse effects of additional FLOPs  are somewhat compensated by the reduced complexity of low-rank attention, it only happens when the input sequence is considerably large. For short  input sequences, Linformer effectively increases the complexity by increasing the FLOPs instead of reducing them - demonstrated in the Experiments section of this work.  Our  method directly generates low-rank \textit{\underbar{K}} and \textit{\underbar{V}} embeddings from the input feature map \textit{X} in a learnable instance-guided manner. This reduces  parameter count as well as  FLOPs due to skipping of the full-sized \textit{K} and \textit{V} generation. 
For clearly establishing the technical difference between GLMHA and Linformer, we illustrate their internal mechanisms side-by-side in \cref{fig_4}.

\noindent{\bf Plug-n-play Enhancement ---}
The proposed GLMHA acts as a stand-alone plug-n-play module for enhancing the existing techniques that employ  CSA. The sequence length independence of GLMHA for different hyperparameter settings  makes it flexible for plugging it in at different depths of the neural network. We extend three SOTA  models, namely MST-L \cite{r42}, MST++ \cite{r43}, and Restormer \cite{r1} with our proposed GLMHA on respective four different image restoration and spectral reconstruction tasks. Note that, these enhancements are an intended contribution of this work. Besides, we also extend these neural networks using related existing complexity reduction techniques for bench-marking with the proposed GLMHA. We provide details of these enhancements in the next section.


\vspace{-2mm}

\section{Experiments}\label{sec:experiments}
\vspace{-2mm}
To validate the effectiveness of GLMHA, we experiment with the channel self-attention (CSA)  of three SOTA models MST \cite{r42}, MST++ \cite{r43}, and Restormer \cite{r1}. We alter these networks by replacing their CSA with the proposed GLMHA. Moreover, we also carefully apply the existing complexity reduction techniques to these methods for benchmarking. Owing to  their high relevance to the considered tasks and popularity, we select Reformer \cite{r8}, Linformer \cite{r2}, and Performer \cite{r9} as the existing complexity reduction techniques. Our experiments are conducted for Hyperspectral Image (HSI) reconstruction from RBG images \cite{akhtar2018hyperspectral}, HSI reconstruction from Snapshot Compressive Imaging  \cite{r42}, Image Deraining \cite{r1}, and Single Image Motion Deblurring \cite{r1}. Additionally, we also perform ablation studies on the deraining task. {We used the same experimental settings for training and testing the models as used by SOTA models.} In the reported results,  default CSA usage by the existing methods is provided as a reference in grey cells of the Tables. Boldfaced values in green cells show the best performing complexity reduction results. 

\noindent\textbf{HSI reconstruction from RGB Images ---}\label{hsi-rgb} We conduct HSI reconstruction from RGB images using the SOTA model MST++~\cite{r43}. We use the NTIRE 2022 Spectral Reconstruction Challenge \cite{r43} dataset to train the models. It contains 1,000 RGB-HSI image pairs, split into 950 train, 50 validation, and 50 test images. Each image is of size 482$\times$512 and has 31 channels in the wavelength range of 400-700~nm. 

\begin{figure}[t]
\begin{center}
\includegraphics[scale =0.51]{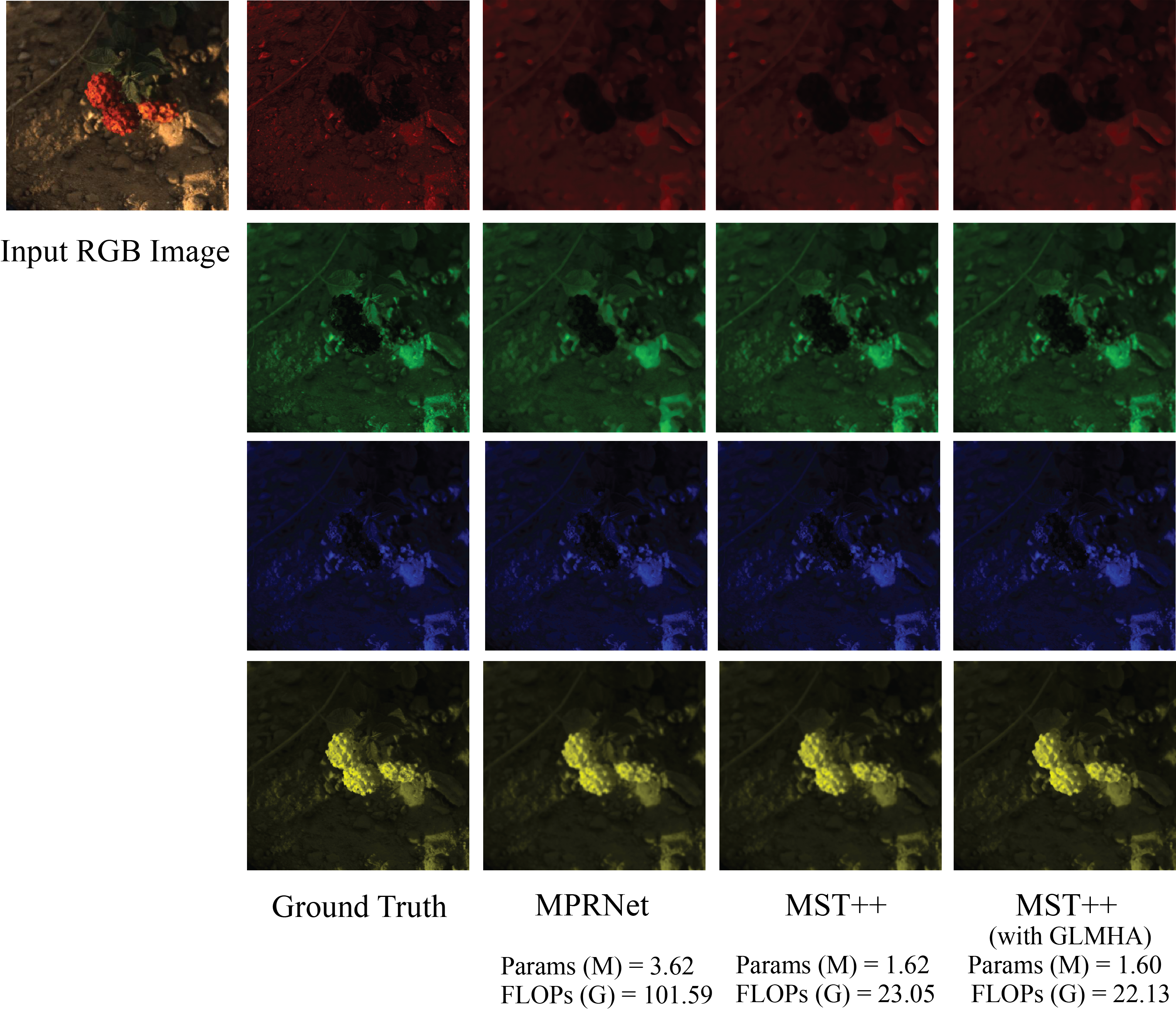}
\end{center}

   \caption{Representative qualitative results of HSI reconstruction. Four out of 31 generated spectral images are shown for each model. Best viewed enlarged. }
\label{fig_5}

\end{figure}

\cref{tab1} summarises the results of our experiments. 
It is emphasized that 
MST++ is already a lightweight architecture and the winner of the NTIRE 2022 challenge on Spectral Recovery. It has  the best performance in terms of least FLOPs and parameter count, and the best PSNR values. It consists of cascaded U-shaped shallow encoder-decoder networks with skip connections. Each encoder-decoder network is three levels deep with CSA used in all transformer blocks at each level. 
Our enhancement replaces CSA with GLMHA. 
Clearly from  \cref{tab1},  existing techniques do not  perform well as compared to our GLMHA due to their strong dependence on the large-sized sequences to be effective. 
Notice that the existing methods in \cref{tab1} actually increase the FLOPs  without parameter reduction. Moreover, they end-up significantly affecting the PNSR values. 
This is because the complexity reduction of these methods fail to overcome their computational overhead for relatively shorter sequences, which are relevant for this task. Furthermore, these methods do not account for parameter reduction. GLMHA resolves these problems. This allows it to achieve a significant FLOPs reduction and $\sim$20K less parameters while closely maintaining the original PNSR. We also show representative qualitative results in \cref{fig_5}. 

\begin{table}[t!]
\begin{center}
\begin{tabular}{|lccc|}
\hline
\multicolumn{4}{|c|}{\textbf{HSI reconstruction from RGB Images}}                                                                           \\ \hline
\multicolumn{1}{|l|}{\textbf{Method}} & \multicolumn{1}{c|}{\textbf{FLOPs (G)}} & \multicolumn{1}{c|}{\textbf{Params (M)}} & \textbf{PSNR}  \\ \hline
\rowcolor{Gray}
\multicolumn{1}{|l|}{CSA (Default)}         & \multicolumn{1}{c|}{23.05}              & \multicolumn{1}{c|}{1.62}                & 34.32          \\
\multicolumn{1}{|l|}{Reformer}         & \multicolumn{1}{c|}{23.17 (+0.53$\%$)}              & \multicolumn{1}{c|}{1.62 (-0.00\%)}                & 33.71          \\
\multicolumn{1}{|l|}{Linformer}         & \multicolumn{1}{c|}{23.17 (+0.56\%)}               & \multicolumn{1}{c|}{1.62 (-0.00\%)}                &  33.79           \\
\multicolumn{1}{|l|}{Performer}         & \multicolumn{1}{c|}{23.15 (+0.47$\%$)}              & \multicolumn{1}{c|}{1.62 (-0.00$\%$)}                & 33.57          \\
\rowcolor{lightgreen}
\multicolumn{1}{|l|}{\textbf{GLMHA (Our)}}         & \multicolumn{1}{c|}{\textbf{22.13 (-3.95\%)}}              & \multicolumn{1}{c|}{\textbf{ 1.60 (-1.24\%)}}                & \textbf{34.21}            \\
\hline
\end{tabular}
\end{center}
\caption{NTIRE 2022 Spectral Reconstruction Challenge performance. CSA (Default) results in grey cells are provided for reference only. GLMHA strongly outperforms the existing methods in both FLOPs and parameter count reduction, while closely maintaining the original PNSR of CSA.}
\label{tab1}
\end{table}

\begin{figure}[t!]
\begin{center}
\includegraphics[scale =0.5]{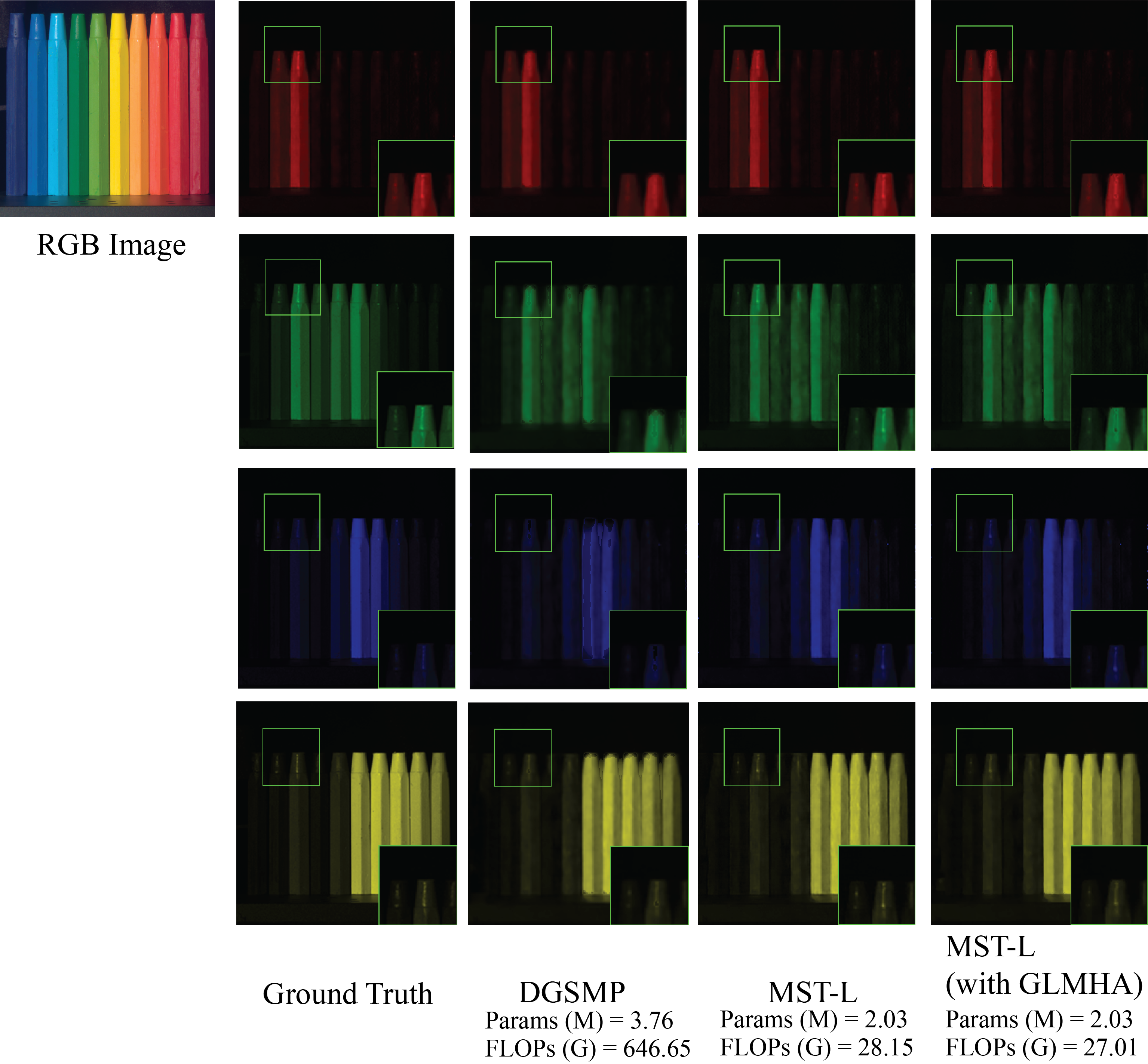}
\end{center}

   \caption{Representative qualitative results for HSI Reconstruction from SCI. Boxed regions are magnified at the bottom right of each image. }
\label{fig_6}
\end{figure}

\noindent\textbf{HSI reconstruction from Snapshot Compressive Imaging (SCI) ---}  For this task, MST-L \cite{r42} is used as the base model, which instead of using cascaded shallow encoder-decoder networks, consists of a single deep encoder-decoder network with skip connections. In MST-L, all the transformer blocks use CSA. We test the effectiveness of the complexity reduction techniques using the same approach as above. Although MST-L  uses more attention blocks,  the number of channels in each block is still not high enough for the methods like Linformer, Performer, and Reformer to be effective. \cref{tab2} shows that our GLMHA again outperform these approaches considerably. We also provide representative qualitative results in \cref{fig_6}. 
\begin{table}[t!]
\begin{center}
\begin{tabular}{|lccc|}
\hline
\multicolumn{4}{|c|}{\textbf{HSI reconstruction in SCI}}                                                                           \\ \hline
\multicolumn{1}{|l|}{\textbf{Method}} & \multicolumn{1}{c|}{\textbf{FLOPs (G)}} & \multicolumn{1}{c|}{\textbf{Params (M)}} & \textbf{PSNR}  \\ \hline
\rowcolor{Gray}
\multicolumn{1}{|l|}{CSA (Default)}         & \multicolumn{1}{c|}{28.15}              & \multicolumn{1}{c|}{2.03}                & 35.18          \\
\multicolumn{1}{|l|}{Reformer}         & \multicolumn{1}{c|}{ 28.37 (+0.79$\%$)}              & \multicolumn{1}{c|}{2.03 (-0.00\%)}                &    34.11       \\
\multicolumn{1}{|l|}{Linformer}         & \multicolumn{1}{c|}{28.27 (+0.42\%)}               & \multicolumn{1}{c|}{2.03 (-0.00\%)}                & 34.67            \\
\multicolumn{1}{|l|}{Performer}         & \multicolumn{1}{c|}{ 28.34 (+0.68$\%$)}              & \multicolumn{1}{c|}{2.03 (-0.00$\%$)}                & 34.21          \\
\rowcolor{lightgreen}
\multicolumn{1}{|l|}{\textbf{GLMHA (Our)}}         & \multicolumn{1}{c|}{\textbf{27.01 (-4.05\%)}}              & \multicolumn{1}{c|}{\textbf{2.00 (-1.33\%)}}                & \textbf{35.06}            \\
\hline
\end{tabular}
\end{center}
\caption{Comparative performance analysis for the use of different complexity reduction techniques to make CSA lightweight on the task of Snapshot Compressive Imaging.}
\label{tab2}
\end{table}

\begin{table}[t!]
\begin{center}
\begin{tabular}{|lccc|}
\hline
\multicolumn{4}{|c|}{\textbf{Single Image Motion Deblurring}}                                                                           \\ \hline
\multicolumn{1}{|l|}{\textbf{Self Attention}} & \multicolumn{1}{c|}{\textbf{FLOPs (G)}} & \multicolumn{1}{c|}{\textbf{Params (M)}} & \textbf{PSNR}  \\ \hline
\rowcolor{Gray}
\multicolumn{1}{|l|}{CSA (Default)}         & \multicolumn{1}{c|}{140.99}              & \multicolumn{1}{c|}{26.13}                & 32.88          \\
\multicolumn{1}{|l|}{Reformer}         & \multicolumn{1}{c|}{ 137.12 (-2.74$\%$)}              & \multicolumn{1}{c|}{26.13 (-0.00\%)}                &    32.65       \\
\multicolumn{1}{|l|}{Linformer}         & \multicolumn{1}{c|}{137.26 (-2.64$\%$)}               & \multicolumn{1}{c|}{26.13 (-0.00\%)}                & 32.67            \\
\multicolumn{1}{|l|}{Performer}         & \multicolumn{1}{c|}{138.26 (-1.94\%)}              & \multicolumn{1}{c|}{26.13 (-0.00$\%$)}                & 30.75          \\
\rowcolor{lightgreen}
\multicolumn{1}{|l|}{\textbf{GLMHA (Our)}}         & \multicolumn{1}{c|}{\textbf{133.27 (-5.48\%)}}              & \multicolumn{1}{c|}{\textbf{25.76 (-1.40\%)}}                & \textbf{32.77}            \\
\hline
\end{tabular}
\end{center}
\caption{Comparative performance analysis for the use of different complexity reduction techniques to make CSA lightweight on the task of Single Image Motion Deblurring.}
\label{tab3}
\end{table}
\begin{table}[t!]
\begin{center}
\begin{tabular}{|lccc|}
\hline
\multicolumn{4}{|c|}{\textbf{Image Deraining}}                                                                           \\ \hline
\multicolumn{1}{|l|}{\textbf{Self Attention}} & \multicolumn{1}{c|}{\textbf{FLOPs (G)}} & \multicolumn{1}{c|}{\textbf{Params (M)}} & \textbf{PSNR}  \\ \hline
\rowcolor{Gray}
\multicolumn{1}{|l|}{CSA (Default)}         & \multicolumn{1}{c|}{140.99}              & \multicolumn{1}{c|}{26.13}                & 33.94          \\
\multicolumn{1}{|l|}{Reformer}         & \multicolumn{1}{c|}{ 138.25 (-1.94$\%$)}              & \multicolumn{1}{c|}{26.13 (-0.00\%)}                &    33.46       \\
\multicolumn{1}{|l|}{Linformer}         & \multicolumn{1}{c|}{137.16 (-2.71$\%$)}               & \multicolumn{1}{c|}{26.13 (-0.00\%)}                & 33.69            \\
\multicolumn{1}{|l|}{Performer}         & \multicolumn{1}{c|}{139.61 (-0.98\%)}              & \multicolumn{1}{c|}{26.13 (-0.00$\%$)}                & 29.55          \\
\rowcolor{lightgreen}
\multicolumn{1}{|l|}{\textbf{GLMHA (Our)}}         & \multicolumn{1}{c|}{\textbf{134.53 (-4.58\%)}}              & \multicolumn{1}{c|}{\textbf{25.78 (-1.34\%)}}                & \textbf{33.86}            \\
\hline
\end{tabular}
\end{center}
\caption{Comparative performance analysis for the use of different complexity reduction techniques to make CSA lightweight on the task of Single Image Deraining.}

\label{tab4}

\end{table}

\begin{table*}[t!]
\begin{center}
 \setlength{\tabcolsep}{3.5pt}
\def\pz{\phantom{0}}
\renewcommand{\arraystretch}{1.1}
\resizebox{\textwidth}{!}{%
\begin{tabular}{|llll|llll|}
\hline
\multicolumn{4}{|c|}{\textbf{HSI reconstruction from SCI}}  & \multicolumn{4}{c|}{\textbf{HSI reconstruction from RGB}} \\ \hline
\multicolumn{1}{|c|}{\textbf{Model}}           & \multicolumn{1}{c|}{\textbf{FLOPs }} & \multicolumn{1}{c|}{\textbf{Params}} & \multicolumn{1}{c|}{\textbf{PSNR}} & \multicolumn{1}{c|}{\textbf{Model}}      & \multicolumn{1}{c|}{\textbf{FLOPs}} & \multicolumn{1}{c|}{\textbf{Params }} & \multicolumn{1}{c|}{\textbf{PSNR}} \\ 
\multicolumn{1}{|c|}{}  & \multicolumn{1}{c|}{\textbf{(G)}} & \multicolumn{1}{c|}{\textbf{(M)}} &  \multicolumn{1}{c|}{}
 & \multicolumn{1}{c|}{} & \multicolumn{1}{c|}{\textbf{(G)}} & \multicolumn{1}{c|}{\textbf{(M)}} &  \multicolumn{1}{c|}{}  \\ \hline
\multicolumn{1}{|l|}{$\lambda$-Net \cite{r54}} & \multicolumn{1}{r|}{117.98}             & \multicolumn{1}{r|}{62.64}               & 28.53                              & \multicolumn{1}{l|}{MIRNet \cite{r27}}   & \multicolumn{1}{r|}{42.95}              & \multicolumn{1}{r|}{3.75}                & 33.29                              \\
\multicolumn{1}{|l|}{DIP-HSI \cite{r55}}       & \multicolumn{1}{r|}{163.48}             & \multicolumn{1}{r|}{1.19}                & 31.26                              & \multicolumn{1}{l|}{Restormer \cite{r1}} & \multicolumn{1}{r|}{93.77}              & \multicolumn{1}{r|}{15.11}               & 33.40                              \\
\multicolumn{1}{|l|}{TSA-Net \cite{r56}}       & \multicolumn{1}{r|}{110.06}             & \multicolumn{1}{r|}{44.25}               & 31.46                              & \multicolumn{1}{l|}{MPRNet \cite{r27}}   & \multicolumn{1}{r|}{101.59}             & \multicolumn{1}{r|}{3.62}                & 33.50                              \\
\rowcolor{lightskyblue}
\multicolumn{1}{|l|}{DGSMP \cite{r57}}         & \multicolumn{1}{r|}{646.65}             & \multicolumn{1}{r|}{3.76}                & 32.63                              & \multicolumn{1}{l|}{MST-L \cite{r42}}    & \multicolumn{1}{r|}{32.07}              & \multicolumn{1}{r|}{2.45}                & 33.90                              \\
\rowcolor{Gray}
\multicolumn{1}{|l|}{MST-L \cite{r42}}         & \multicolumn{1}{r|}{28.15}              & \multicolumn{1}{r|}{2.03}                & \textbf{35.18}                     & \multicolumn{1}{l|}{MST++ \cite{r43}}    & \multicolumn{1}{r|}{23.05}              & \multicolumn{1}{r|}{1.62}                & \textbf{34.32}                              \\
\rowcolor{lightgreen}
\multicolumn{1}{|l|}{MST-L (with GLMHA)}                & \multicolumn{1}{r|}{\textbf{27.01}}              & \multicolumn{1}{r|}{\textbf{2.00}}                & 35.06                              & \multicolumn{1}{l|}{MST++ (with GLMHA)}           & \multicolumn{1}{r|}{\textbf{22.13}}              & \multicolumn{1}{r|}{\textbf{1.60}}                & 34.21                              \\ \hline
\end{tabular}}
\end{center}
\caption{SOTA benchmarking for HSI reconstruction from SCI and RGB. GLMHA-enhanced models (in green cells) retain higher PSNR for each task as compared to the second-best performers (blue cells), while achieving considerable FLOPs and Parameter reduction over the original models (grey cells). Best performance is boldfaced.}
\label{tab5_1}
\end{table*}

\begin{table*}[t!]
\begin{center}
 \setlength{\tabcolsep}{3.5pt}
\def\pz{\phantom{0}}
\renewcommand{\arraystretch}{1.1}
\resizebox{\textwidth}{!}{%
\begin{tabular}{|llll|llll|}
\hline
\multicolumn{4}{|c|}{\textbf{Single Image Deblurring}}  & \multicolumn{4}{c|}{\textbf{Deraining}}                                                                                                                          \\ \hline
\multicolumn{1}{|c|}{\textbf{Model}}           & \multicolumn{1}{c|}{\textbf{FLOPs }} & \multicolumn{1}{c|}{\textbf{Params}} & \multicolumn{1}{c|}{\textbf{PSNR}} & \multicolumn{1}{c|}{\textbf{Model}}      & \multicolumn{1}{c|}{\textbf{FLOPs}} & \multicolumn{1}{c|}{\textbf{Params }} & \multicolumn{1}{c|}{\textbf{PSNR}} \\ 
\multicolumn{1}{|c|}{}  & \multicolumn{1}{c|}{\textbf{(G)}} & \multicolumn{1}{c|}{\textbf{(M)}} &  \multicolumn{1}{c|}{} & \multicolumn{1}{c|}{} & \multicolumn{1}{c|}{\textbf{(G)}} & \multicolumn{1}{c|}{\textbf{(M)}} &   \\ \hline
\multicolumn{1}{|l|}{DBGAN \cite{r53}}      & \multicolumn{1}{r|}{759.00}             & \multicolumn{1}{r|}{110.00}              & 31.10                              & \multicolumn{1}{l|}{PreNet \cite{r48}}   & \multicolumn{1}{r|}{66.00}              & \multicolumn{1}{r|}{100.00}              & 29.42                              \\
\multicolumn{1}{|l|}{MIMO-UNet+ \cite{r52}} & \multicolumn{1}{r|}{154.00}             & \multicolumn{1}{r|}{158.00}              & 32.45                              & \multicolumn{1}{l|}{RESCAN \cite{r50}}   & \multicolumn{1}{r|}{32.00}              & \multicolumn{1}{r|}{28.00}               & 28.59                              \\
\multicolumn{1}{|l|}{IPT \cite{r51}}        & \multicolumn{1}{r|}{379.00}             & \multicolumn{1}{r|}{114.00}              & 32.52                              & \multicolumn{1}{l|}{MPRNet \cite{r27}}   & \multicolumn{1}{r|}{141.00}             & \multicolumn{1}{r|}{20.10}               & 32.73                              \\
\rowcolor{lightskyblue}
\multicolumn{1}{|l|}{MPRNet \cite{r27}}     & \multicolumn{1}{r|}{760.00}             & \multicolumn{1}{r|}{20.10}               & 32.66                              & \multicolumn{1}{l|}{MSPFN \cite{r49}}    & \multicolumn{1}{r|}{605.00}             & \multicolumn{1}{r|}{33.50}               & 30.72\\
\rowcolor{Gray}
\multicolumn{1}{|l|}{Restormer \cite{r1}}   & \multicolumn{1}{r|}{140.99}             & \multicolumn{1}{r|}{26.13}               & \textbf{32.92}                              & \multicolumn{1}{l|}{Restormer \cite{r1}} & \multicolumn{1}{r|}{140.99}             & \multicolumn{1}{r|}{26.13}               & \textbf{33.94}                              \\
\rowcolor{lightgreen}
\multicolumn{1}{|l|}{Restormer (with GLMHA)}             & \multicolumn{1}{r|}{\textbf{133.27}}             & \multicolumn{1}{r|}{\textbf{25.76}}               & 32.77                              & \multicolumn{1}{l|}{Restormer (with GLMHA)}           & \multicolumn{1}{r|}{\textbf{134.53}}             & \multicolumn{1}{r|}{\textbf{25.78}}               & 33.86                              \\ \hline
\end{tabular}}
\end{center}
\caption{SOTA benchmarking for Single Image Deblurring and Deraining. GLMHA enhanced models (in green cells) retain higher PSNR for each task as compared to the second-best performers (blue cells), while achieving considerable FLOPs and Parameter reduction over the original models (grey cells). Best performance is boldfaced.}
\label{tab5_2}
\end{table*}

\begin{figure}
\begin{center}
\includegraphics[scale =0.65]{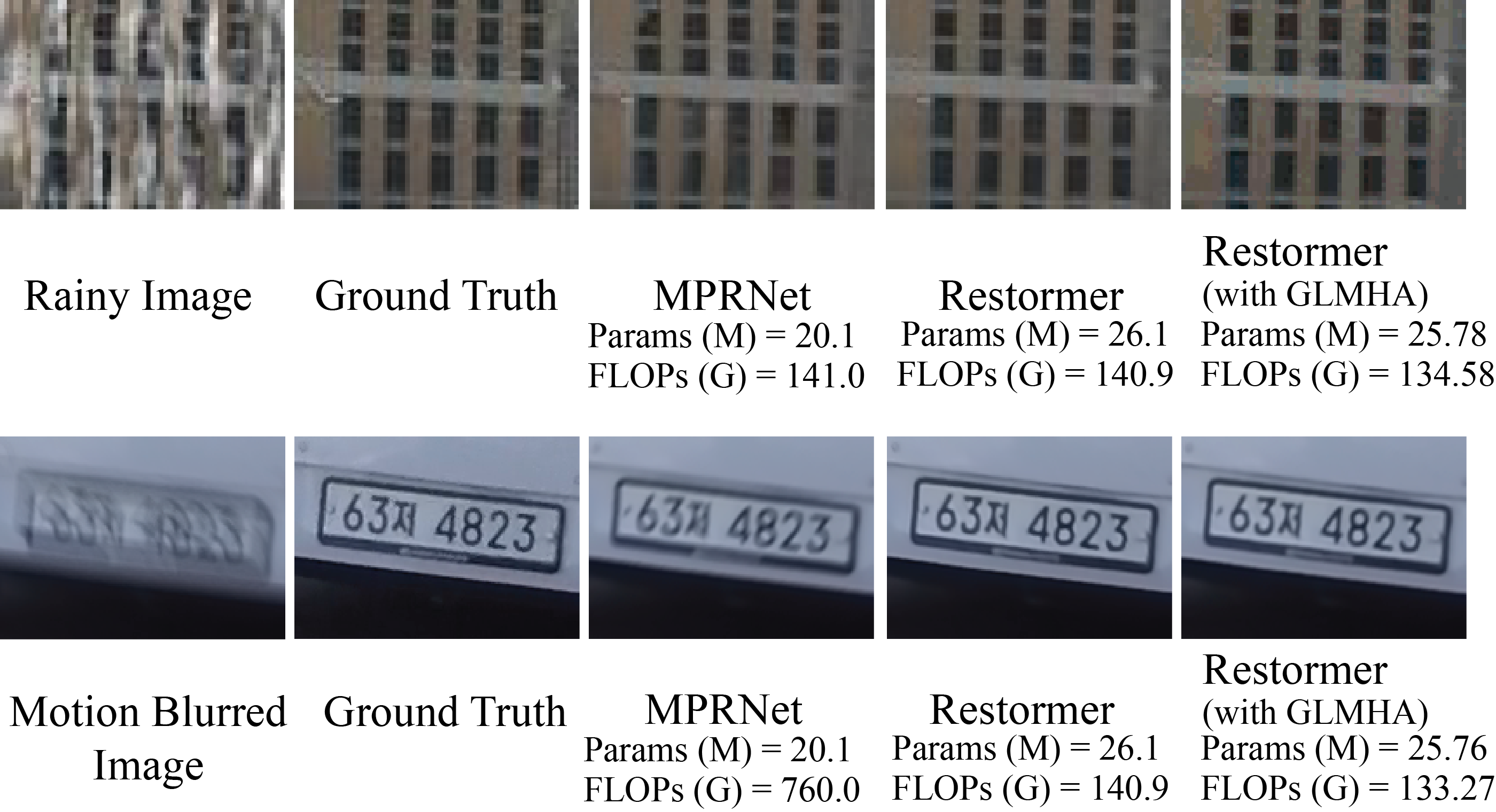}
\end{center}

   \caption{Qualitative Comparison of our work with other models on the task of Image Deraining (Top Row), and on Image Deblurring (Bottom Row).}
\label{fig_7}

\end{figure}

\noindent\textbf{Single Image Motion Deblurring ---}  We perform experiments for this task using Restormer \cite{r1}, which is a four-level deep encoder-decoder architecture with skip connections. Like MST++ and MST-L, it also uses CSA in its  transformer blocks, albeit with more channels. The number of channels doubles going deeper into the network, similar to MST++ and MST-L. Following \cite{r1}, we use the GoPro dataset \cite{r1} for  training and testing in our experiments. This dataset comprises 2,103 images for training and 1,111 images for testing. The image size is 1280$\times$720. Since this architecture is deeper and uses more channels the Reformer, Linformer, and Performer are able to show better complexity reduction performance on Restormer (as compared to MST++ and MST-L). However, as evident from \cref{tab3}, GLMHA still outperforms these methods by a large margin in FLOPs reduction, while also using 370K less parameters as compared to the original model. Again, GLMHA is able to closely retain the original PNSR value as well. Representative qualitative results are shown in ~\cref{fig_7}(top).

\noindent\textbf{Image Deraining ---}   For this task,  Restormer \cite{r1} shows the SOTA performance. So, we choose Restormer as our base model. We employ Rain13k dataset \cite{r1}  for training, and follow the exact evaluation protocol as in~\cite{r1}. We test the models on 5 deraining datasets, i.e., Rain100H, Rain100L, Test100, Test1200, and Test2800, and summarize the average PSNR values in \cref{tab4}. For this task, GLMHA is able to reduce nearly 6.5G FLOPs and 350K parameters for the original model while retaining close average PNSR value. Representative qualitative results for this task are shown in \cref{fig_7}(bottom).

\vspace{-5mm}
\noindent\textbf{SOTA benchmarking ---} Whereas \cref{tab1} to \cref{tab4} affirm that GLMHA shows the best complexity reduction efficacy for image restoration and spectral reconstruction tasks, in \cref{tab5_1} and \cref{tab5_2} we benchmark the performance of GLMHA-boosted models against the other existing methods for the tasks. From the tables, it is clear that all the GLMHA-boosted models can retain their PNSR values to stay abreast of the second-best performing techniques for the tasks. On the other hand, each method significantly benefits from GLMHA enhancements in terms of efficiency. 

\section{Ablation Study}
We present three different experiments with Restormer \cite{r1} for image deraining as an ablation study.

\noindent\textbf{Effect of changing size reduction ratio for \textit{K} and \textit{V} ---} We trained the network using different size reduction ratios for \textit{K} and \textit{V} embeddings in GLMHA attention, keeping the $\alpha$ value to 0.60 (best performer). It is found that size reduction of \textit{K} and \textit{V} beyond half the original size can cause a sharp decrease in performance, accompanied with a proportional FLOPs reduction. The results are summarized in \cref{tab6}.



\noindent\textbf{Effect of changing scaling factor  ---} We changed the  $\alpha$ value while keeping the size reduction ratio to 0.5. The best performance was achieved when the $\alpha$ value was set to 0.6. \cref{tab7} shows the results of this experiment. The best $\alpha$ value was also consistent in experiments on other datasets.\par

\noindent\textbf{Effect of incorporating GLMHA at different depth levels of the network ---} We progressively replaced CSA with our proposed GLMHA at different depths of the network. The use of GLMHA at deeper layers has a more prominent effect on FLOPs and parameter reduction due to a larger number of channels for self-attention. \cref{tab8} shows the results of this experiment.

\begin{table}[t]
\begin{center}

\begin{tabular}{|c|c|c|c|c|}
\hline
\textbf{Method} & \textbf{\begin{tabular}[c]{@{}c@{}}Size reduction ($\%$)\\ \textit{K} and \textit{V}\end{tabular}} & \textbf{\begin{tabular}[c]{@{}c@{}}FLOPs \\ (G)\end{tabular}} & \textbf{\begin{tabular}[c]{@{}c@{}}Params\\ (M)\end{tabular}} & \textbf{PSNR} \\ \hline
CSA             & -                                                                       & 140.99                                                        & 26.13                                                         & 33.94         \\
GLMHA           & 25                                                                       & 139.45                                                        & 26.05                                                         & 33.88         \\
GLMHA           & 50                                                                       & 134.53                                                        & 25.78                                                         & 33.86         \\
GLMHA           & 75                                                                      & 126.11                                                        & 24.87                                                         & 33.10         \\
GLMHA           & 90                                                                      & 124.79                                                        & 23.31                                                         & 29.45         \\ \hline
\end{tabular}
\end{center}
\caption{Effect of changing \textit{K} and \textit{V} size reduction ratio. Restormer model for deraining task is used with $\alpha =0.6$.}
\label{tab6}
\end{table}
\begin{table}[t]
    \centering
    \begin{tabular}{|c|c|c|c|c|c|} \hline
        $\alpha$ & 0.3 & 0.4 & 0.5 & 0.6 & 0.7  \\ \hline
         \textbf{PNSR} & 33.80 & 33.80 & 33.81 & 33.86 & 33.74 \\ \hline
    \end{tabular}
    \caption{Effect of changing the scaling factor $\alpha$ at \textit{K} and \textit{V} size reduction of 50\%.}
    \label{tab7}
\end{table}
\begin{table}[t!]
\begin{center}

\begin{tabular}{|cccc|c|c|c|}
\hline
\multicolumn{4}{|c|}{\textbf{Level}}                                                                                                         & \multirow{2}{*}{\textbf{\begin{tabular}[c]{@{}c@{}}FLOPs \\ (G)\end{tabular}}} & \multirow{2}{*}{\textbf{\begin{tabular}[c]{@{}c@{}}Params \\ (M)\end{tabular}}} & \multirow{2}{*}{\textbf{PSNR}} \\ \cline{1-4}
\multicolumn{1}{|c|}{\textbf{Level 1}} & \multicolumn{1}{c|}{\textbf{Level 2}} & \multicolumn{1}{c|}{\textbf{Level 3}} & \textbf{Bottleneck} &                                                                                &                                                                                 &                                \\ \hline
\multicolumn{1}{|c|}{CSA}              & \multicolumn{1}{c|}{CSA}              & \multicolumn{1}{c|}{CSA}              & CSA                 & 140.99                                                                         & 26.12                                                                           & 33.94                          \\
\multicolumn{1}{|c|}{GLMHA}            & \multicolumn{1}{c|}{CSA}              & \multicolumn{1}{c|}{CSA}              & CSA                 & 139.98 (-0.77$\%$)                                                                & 26.10 (-0.27$\%$)                                                                  & 33.90                          \\
\multicolumn{1}{|c|}{GLMHA}            & \multicolumn{1}{c|}{GLMHA}            & \multicolumn{1}{c|}{CSA}              & CSA                 & 138.80 (-1.55$\%$)                                                                & 26.05 (-0.27$\%$)                                                                  & 33.89                          \\
\multicolumn{1}{|c|}{GLMHA}            & \multicolumn{1}{c|}{GLMHA}            & \multicolumn{1}{c|}{GLMHA}            & CSA                 & 137.40 (-2.55$\%$)                                                                & 25.93 (-0.75$\%$)                                                                  & 33.87                          \\
\multicolumn{1}{|c|}{GLMHA}            & \multicolumn{1}{c|}{GLMHA}            & \multicolumn{1}{c|}{GLMHA}            & GLMHA               & 134.53 (-4.58$\%$)                                                                & 25.78 (-1.34$\%$)                                                                  & 33.86                          \\ \hline
\end{tabular}
\end{center}
\caption{Effect of incorporating GLMHA at different depths of Restormer. Computational progressively improves.}
\label{tab8}
\vspace{-8mm}
\end{table}



\section{Conclusion}
We proposed an efficient self-attention mechanism GLMHA that generates low-rank embeddings for self-attention in an instance-guided manner. This makes low-rank embeddings rich with more relevant information. Overall, it reduces parameter and FLOPs count, and prevents a significant drop in performance. The proposed GLMHA is targeted to replace CSA commonly used in the  SOTA models for image restoration and spectral reconstruction tasks. 
Our extensive experiments with three models for four different tasks establish that the proposed technique can significantly reduce the computational cost of the models while closely retaining the original performance. 

\bibliographystyle{splncs04}
\bibliography{main}
\end{document}